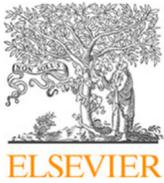
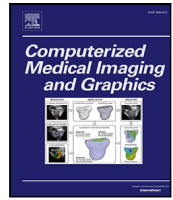

# DuetMatch: Harmonizing semi-supervised brain MRI segmentation via decoupled branch optimization

Thanh-Huy Nguyen [a],[1], Hoang-Thien Nguyen [b],[1], Vi Vu [c], Ba-Thinh Lam [b], Phat Huynh [b], Tianyang Wang [d], Xingjian Li [a], Ulas Bagci [e], Min Xu [a],*

[a] *Carnegie Mellon University, Pittsburgh, 15213, PA, USA*
[b] *PASSIO Lab, North Carolina A&T State University, Greensboro, 27411, NC, USA*
[c] *Ho Chi Minh University of Technology, Ho Chi Minh, 70000, Viet Nam*
[d] *University of Alabama at Birmingham, Birmingham, 35294, AL, USA*
[e] *Northwestern University, Evanston, 60208, IL, USA*

A B S T R A C T

The limited availability of annotated data in medical imaging makes semi-supervised learning increasingly appealing for its ability to learn from imperfect supervision. Recently, teacher-student frameworks have gained popularity for their training benefits and robust performance. However, jointly optimizing the entire network can hinder convergence and stability, especially in challenging scenarios. To address this for medical image segmentation, we propose *DuetMatch*, a novel dual-branch semi-supervised framework with asynchronous optimization, where each branch optimizes either the encoder or decoder while keeping the other frozen. To improve consistency under noisy conditions, we introduce **Decoupled Dropout Perturbation**, enforcing regularization across branches. We also design **Pairwise CutMix Cross-Guidance** to enhance model diversity by exchanging pseudo-labels through augmented input pairs. To mitigate confirmation bias from noisy pseudo-labels, we propose **Consistency Matching**, refining labels using stable predictions from frozen teacher models. Extensive experiments on benchmark brain MRI segmentation datasets, including ISLES2022 and BraTS, show that DuetMatch consistently outperforms state-of-the-art methods, demonstrating its effectiveness and robustness across diverse semi-supervised segmentation scenarios.

## 1. Introduction

Brain tissue and tumor segmentation is essential for accurate diagnosis, anatomical analysis, and understanding of brain disorders, supporting fields like neuroscience, mental health research, and medical imaging (Feigin et al., 2022). For example, brain tumor segmentation aids in diagnosing gliomas, which account for over half of all primary central nervous system tumors and cause many deaths annually (Zhou, 2024). Stroke segmentation also plays a key role by identifying affected regions and tissue damage, guiding clinical decisions and treatment planning. MRI is the preferred imaging modality due to its high contrast between brain tissues, revealing both structural and pathological changes.

Numerous deep learning methods have been developed for brain segmentation (Nguyen et al., 2025a), with Convolutional Neural Networks (CNNs), such as U-Net (Ronneberger et al., 2015) and V-Net (Milletari et al., 2016), being the most prevalent. Recently, general segmentation models such as SAM (Kirillov et al., 2023) and MedSAM (Ma et al., 2024) have gained popularity due to their zero-shot segmentation capability across various medical datasets. However, these models typically depend on large, fully annotated datasets, which are costly and time-consuming to produce. To address this, semi-supervised learning (SSL) has emerged as a promising solution, enabling the use of limited labeled data alongside abundant unlabeled samples to enhance model generalization in medical imaging (Nguyen et al., 2025b).

Semi-supervised semantic segmentation has been predominantly driven by two major paradigms: (1) consistency regularization and (2) co-training strategies. Consistency regularization methods (Zou et al., 2021; Yang et al., 2023) enforce the principle that a model should produce consistent predictions under different input perturbations. On the other hand, co-training approaches (Chen et al., 2021; Fan et al., 2022; Wang et al., 2023; Pan et al., 2025; Vu et al., 2025) aim to exploit complementary information across different views or distinct model branches to enhance learning. In recent years, student-teacher frameworks (Tarvainen and Valpola, 2017; Jin et al., 2022; Bai et al., 2023) have gained significant attention due to their robustness and effective training dynamics. These methods leverage the Exponential Moving Average (EMA) strategy (Tarvainen and Valpola, 2017), where a teacher model, which is maintained as an EMA of the student, is used to generate more stable pseudo-labels, while the student continuously transfers updated knowledge back to the teacher through the EMA update. However, these frameworks typically perform optimization and EMA updates over the entire model, which may be suboptimal in challenging scenarios, such as when the model struggles to converge or when certain samples are particularly difficult to learn from. In such cases, optimizing the full network simultaneously can lead to subpar performance or convergence to poor local minima. Based on this observation, we hypothesize that optimizing individual components of the network (e.g., encoder and decoder) separately can facilitate more effective learning and lead to better generalization by reducing the risk of being trapped in unfavorable optimization states.

To achieve this goal, we propose *DuetMatch*, a dual-branch network with asynchronous optimization objectives, where each branch focuses on learning a specific component (encoder or decoder) while keeping the other component frozen. To improve the robustness of our framework, we introduce Decoupled Dropout Perturbation based on the consistency regularization assumption, encouraging each branch to make stable predictions under input noise. However, relying solely on consistency can limit the learning capacity of the model. Therefore, we incorporate a Pairwise CutMix Cross-Guidance strategy to promote model diversity through co-training. This strategy, however, can introduce noise into the pseudo-labels, leading to confirmation bias (Yang et al., 2022). To address this, we finally propose Consistency Matching, which refines the pseudo-labels by integrating them with a more stable and consistent mask, thereby improving their reliability. In summary, our contributions are as follows:

- DuetMatch introduces a dual-branch semi-supervised framework that leverages asynchronous optimization, enabling independent specialization of the encoder and decoder components.
- Decoupled Dropout Perturbation to improve model robustness by enforcing prediction consistency under stochastic noise.
- Pairwise CutMix Cross-Guidance to enhance diversity and a Consistency Matching mechanism to generate more reliable pseudo-labels and reduce confirmation bias.
- We validate the effectiveness and generalizability of our method across famous and widely used brain MRI segmentation benchmarks, including ISLES2022 and BraTS (BraTS2017, BraTS2018, and BraTS2019).

## 2. Related works

### 2.1. Brain MRI segmentation

Brain tumor segmentation has drawn growing interest due to its importance in identifying tumor characteristics, aiding diagnosis, and guiding treatment planning (Saifullah et al., 2025; Amin et al., 2025; Bengtsson et al., 2025). For instance, Saifullah et al. (2025) combined DeepLabV3Plus with an Xception encoder for accurate tumor segmentation from MRI scans. Amin et al. (2025) proposed a hierarchical pipeline with two-stage, multi-view training to enhance severe tumor detection. Bengtsson et al. (2025) proposed a logic-based segmentation strategy for challenging and pediatric tumor cases, using various backbones such as U-Net, nnU-Net, and Transformers. However, their method relies on detailed annotations of tumors and sub-compartments, a common challenge in supervised segmentation.

Stroke segmentation, another key task in brain imaging, is crucial for diagnosing cerebrovascular diseases, named stroke being the second leading cause of death and third of disability worldwide. Ashtari et al. (2023) proposed a low-rank matrix decomposition method to improve interpretability and reduce computation. Zhang et al. (2022) addressed data scarcity via few-shot learning by transferring knowledge from glioma datasets. Liu et al. (2020) used attention mechanisms to segment ischemic stroke regions, while Wu et al. (2024) combined feature refinement and protection modules to capture detailed local and global features, enhancing performance.

### 2.2. Semi-supervised medical image segmentation

Semi-supervised medical image segmentation (SSMIS) has gained significant attention in recent years (Yu et al., 2019; Li et al., 2020; Wu et al., 2021; Luo et al., 2021, 2022; Wu et al., 2022; Bai et al., 2023; Adiga V. et al., 2024; Pham et al., 2025; Xu et al., 2023; Zhou et al., 2025; Jin et al., 2025, 2024), supporting clinical diagnosis across a wide range of applications. Early work like UA-MT (Yu et al., 2019) introduced Monte Carlo Dropout in a consistency learning framework to promote stable predictions under perturbations. SASS-NET (Li et al., 2020) extended this by enforcing shape-aware consistency through adversarial loss on signed distance maps. MC-Net (Wu et al., 2021) targeted challenging regions with mutual consistency to enforce low-entropy, stable outputs across decoders. URPC (Luo et al., 2022) enhanced scale-awareness by aligning pyramid-level predictions with their average. BCP (Bai et al., 2023) tackled distribution mismatch via a two-stage teacher-student scheme with bidirectional copy-paste. DAE-MT (Adiga V. et al., 2024) improved uncertainty estimation by leveraging global contextual cues from masks. ERSR (Zhou et al., 2025) combined adaptive filtering, ellipse-based pseudo-label refinement, and multi-level consistency regularization to improve fetal head ultrasound segmentation. PG-FaNet (Jin et al., 2024) introduced a pseudo-mask guided feature aggregation network with uncertainty-aware consistency regularization for histopathology image segmentation.

However, these methods typically optimize the network end-to-end, which may hinder convergence and performance on difficult samples. Our work introduces an asynchronous optimization strategy with distinct objectives across components, leveraging dropout-based perturbations and co-training to enhance both consistency and diversity in learning.

## 3. Methodology

### 3.1. Overview

Let the labeled dataset be denoted as $\mathcal{D}_l = (x_l, y_l)$, where $x_l$ represents the input images and $y_l$ is the corresponding ground truth segmentation masks. Similarly, we denote the unlabeled dataset as $\mathcal{D}_u = x_u$, which contains only input images without annotations. In conventional semi-supervised learning paradigms, the total optimization objective combines supervised and unsupervised losses, and is formulated as:

$$\mathcal{L} = \mathcal{L}_{sup} + \lambda \cdot \mathcal{L}_{unsup}. \quad (1)$$

Here, $\mathcal{L}_{sup}$ denotes the loss computed on labeled data, $\mathcal{L}_{unsup}$ represents the loss on unlabeled data (often derived from pseudo-labels or consistency regularization), and the hyperparameter $\lambda$ balances the contribution of the unsupervised loss component relative to the supervised loss.

*DuetMatch* (see Section 3.2) adopts a dual-branch learning approach with distinct optimization objectives, allowing each branch to apply a complementary strategy for better use of unlabeled data. To boost robustness, we introduce Decoupled Dropout Perturbation (Section 3.3), using different dropout settings per branch to promote diverse features and reduce co-adaptation. We also apply Pairwise CutMix Cross-Guidance (Section 3.4), which exchanges mixed semantic content between branches to enhance pseudo-supervision and generalization. Finally, Consistency Matching (Section 3.5) enforces agreement between selected predictions, helping stabilize pseudo-label quality during training (see Fig. 1).

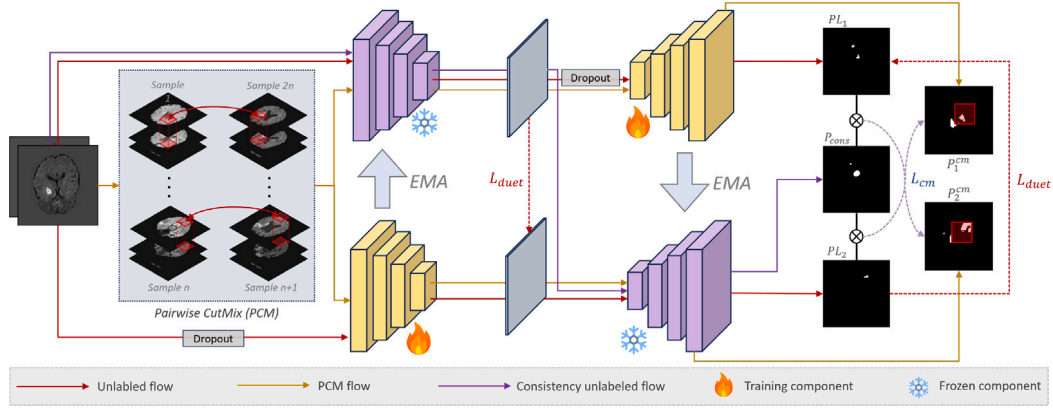

**Fig. 1.** The proposed *DuetMatch* framework is illustrated. First, Decoupled Dropout Perturbation is applied by augmenting the feature map extracted from the first branch and the input sample fed to the second branch using unlabeled data, then the Duet loss is then computed. Second, Pairwise CutMix augmentation is applied to unlabeled inputs. Pseudo-labels are generated independently from both branches and then refined using Consistency Matching with predictions from frozen teacher models. These refined pseudo-labels are exchanged between branches to guide each other, forming the Pairwise CutMix Cross-Guidance mechanism that promotes mutual supervision and enhanced learning diversity.

### 3.2. DuetMatch framework

Let $f$ and $g$ denote the encoder and decoder components of a model within the dual-branch architecture, respectively. We use $\theta_i$ to represent the parameters of the $i$th model branch, where $i \in \{1, 2\}$. Our work is inspired by the design proposed in Pahk et al. (2025), which hypothesized that a model could learn more effectively when each branch focuses on optimizing a specific component. In that framework, a dual-branch network was introduced where each branch freezes either the encoder or decoder and learns the remaining part, using the ground truth mask to guide supervision.

Extending this concept to the semi-supervised learning setting, where ground truth annotations are absent for the unlabeled data, we adapt this decoupling strategy for unlabeled input. However, unlike Pahk et al. (2025), which computes the loss between final predictions and ground truth masks, our approach assumes that the frozen components often yield more stable and generalizable outputs. Therefore, instead of relying solely on the final prediction for supervision, we propose that the learned components should be guided directly by the outputs of the corresponding frozen components in the other branch.

Specifically, let the outputs from each component be defined as follows:

$$F_t = f_{\theta_1^*}(x_u), \qquad P_s = g_{\theta_1}(F_t), \tag{2}$$

$$F_s = f_{\theta_2}(x_u), \qquad P_t = g_{\theta_2^*}(F_s), \tag{3}$$

where $f_{\theta_1^*}$ and $g_{\theta_2^*}$ denote the frozen encoder and decoder from the first branch and the second branch, respectively. The frozen components serve as stable references to supervise the corresponding learned components in the opposite branch.

Assuming that the frozen components generate more consistent and reliable representations, we compute the losses between these reference outputs and those from the components being trained. The overall Duet optimization loss is defined as:

$$\mathcal{L}_{duet} = \alpha \cdot l_F(F_s, F_t) + l_P(P_s, P_t), \tag{4}$$

where $l_F$ represents the feature-level loss between encoder outputs, $l_P$ denotes the segmentation mask loss between decoder predictions and $\alpha$ is a loss weight for controlling the effect on feature representation. In our work, the Mean Square Error is employed as the feature-level loss and the Cross Entropy is used for $l_P$, the $\alpha$ is set at 0.5. This formulation ensures targeted, component-specific supervision, allowing each branch to improve its weaker parts through consistent and guided learning.

For the labeled data, a combination of Cross-Entropy loss and Dice loss was adopted to supervise the training. The predictions from both branches are obtained by combining one learned and one frozen component in each path:

$$P_1^l = g_{\theta_1}(f_{\theta^1}(x_l)), \qquad P_2^l = g^{\theta^2}(f\theta_2(x_l)). \tag{5}$$

The supervised loss for each prediction is defined as a weighted sum of the Cross-Entropy (CE) and Dice loss:

$$l_{sup}(P, y) = CE(P, y) + DICE(P, y). \tag{6}$$

Then, the overall supervised loss is computed by averaging the losses from both prediction paths:

$$\mathcal{L}_{sup} = 0.5 \cdot l_{sup}(P_1^l, y) + 0.5 \cdot l_{sup}(P_2^l, y). \tag{7}$$

### 3.3. Decoupled dropout perturbation

To enhance the robustness and consistency of learning within each branch, we adopt a dropout-based perturbation strategy, inspired by previous works such as Ouali et al. (2020) and Yang et al. (2023). In our method, uniquely, dropout is applied to both the input image and intermediate features as a form of stochastic noise injection. This perturbation encourages the model to produce stable and consistent predictions under varying input conditions, thereby improving generalization and resilience to noise.

By introducing dropout during the forward pass, the output computation can be reformulated as:

$$F_t = f_{\theta_1^*}(x_u), \tag{8}$$

$$F_s^{dropout} = f_{\theta_2}(\text{Dropout}(x_u)), \tag{9}$$

$$P_s^{dropout} = g_{\theta_1}(\text{Dropout}(F_t)), \tag{10}$$

$$P_t = g_{\theta_2^*}(f_{\theta_2}(x_u)), \tag{11}$$

where $F_t$ and $P_t$ are obtained from frozen components without dropout, serving as reliable references, while $F_s$ and $P_s$ are computed using perturbed inputs and intermediate features to simulate uncertainty during learning.

The corresponding loss function incorporating dropout is then defined as:

$$\mathcal{L}_{duet}^{dropout} = \alpha \cdot l_F(F_s^{dropout}, F_t) + l_P(P_s^{dropout}, P_t). \tag{12}$$

This dropout-augmented objective promotes feature-level and prediction-level consistency under input perturbations, leading to more robust semi-supervised learning.

*3.4. Pairwise CutMix cross-guidance*

While the duet strategy encourages modular specialization within each branch, relying solely on it may constrain the learning capacity and diversity of the overall framework. To further enhance the representational diversity and encourage complementary learning across branches, we adopt the co-training assumption, which leverages mutual guidance between distinct models. In particular, we introduce a Pairwise CutMix Cross-Guidance strategy to augment the unlabeled data and enforce collaborative supervision.

First, we apply Pairwise CutMix augmentation to the unlabeled input batch $x_u$. This augmentation is performed between each sample in the batch and its corresponding sample in the reversed batch (i.e., paired with a different sample), ensuring that augmented inputs are generated from distinct sample pairs:

$$x_u^{cm} = x_u \cdot \mathcal{M} + \overline{x}_u \cdot (1 - \mathcal{M}), \tag{13}$$

where $\mathcal{M} \in (0,1)^{W \times H \times D}$ is a randomly generated binary mask controlling the CutMix region, and $\overline{x}_u$ denotes the reversed version of the unlabeled batch.

This CutMix-augmented input $x_u^{cm}$ is then passed through both branches of the network to obtain predictions:

$$P_1^{cm} = g_{\theta_1}(f^{\theta_1}(x_u^{cm})), \qquad P_2^{cm} = g\theta^2(f\theta_2(x_u^{cm})). \tag{14}$$

To generate the pseudo-labels corresponding to these augmented predictions, we compute the predictions on the original (non-augmented) unlabeled input $x_u$:

$$P_1^u = g_{\theta_1^*}(f_{\theta_1^*}(x_u)), \qquad P_2^u = g_{\theta_2^*}(f_{\theta_2}(x_u)) \tag{15}$$

The corresponding pseudo-labels are then constructed by applying the same CutMix mask to the predictions and their reversed counterparts:

$$PL_1 = P_1^u \cdot \mathcal{M} + \overline{P_1^u} \cdot (1 - \mathcal{M}), \tag{16}$$

$$PL_2 = P_2^u \cdot \mathcal{M} + \overline{P_2^u} \cdot (1 - \mathcal{M}), \tag{17}$$

where $\overline{P_i^u}$ denotes the reversed prediction batch from the $i$th branch.

Finally, to promote mutual supervision and model diversity, we apply Cross-Guidance, where each model learns from the pseudo-labels generated by the other branch:

$$\mathcal{L}_{cm} = l_P(P_1^{cm}, PL_2) + l_P(P_2^{cm}, PL_1). \tag{18}$$

This strategy enables effective knowledge transfer between two independently regularized networks, thereby encouraging diverse feature learning and more robust pseudo-labeling under a semi-supervised setting.

*3.5. Consistency matching*

Directly using the raw pseudo-labels generated from each branch for supervision likely introduces significant noise into the training process, potentially leading to suboptimal optimization. To address this issue, we draw inspiration from Tarvainen and Valpola (2017), which demonstrates that predictions from a temporally averaged (or frozen) model are typically more stable, consistent, and reliable.

Motivated by this, we first generate a consistency mask by connecting the frozen encoder and decoder from the two branches:

$$P_{cons} = g_{\theta_2^*}(f_{\theta_1^*}(x_u)). \tag{19}$$

This mask, derived from the most stable components of both branches, is assumed to better reflect the true underlying structure in the unlabeled input.

To enhance the quality of the pseudo-labels used in the Cross-Guidance process, we refine them by element-wise multiplication with

**Table 1**
Comparison of different methods on the FLAIR domain of BraTS 2017.

| | Method | DC (%) | JC (%) | 95HD (%) | ASD (%) |
|---|---|---|---|---|---|
| 100% labeled | V-Net | 83.05 | 73.17 | 8.77 | 3.06 |
| Zero-shot | SAM | 36.36 | 23.39 | 44.39 | 17.07 |
| | MedSAM | 44.97 | 30.21 | 16.95 | 7.12 |
| 10% labeled (20/200) | V-Net | 78.87 | 67.78 | 10.62 | 2.97 |
| | UA-MT (MICCAI'19) | 75.69 | 65.13 | 11.72 | 2.45 |
| | SASSNet (MICCAI'20) | 77.11 | 68.02 | 12.73 | 5.19 |
| | DTC (AAAI'21) | 77.20 | 67.31 | 12.25 | 3.04 |
| | MC-Net (MICCAI'21) | 81.07 | 72.05 | 9.63 | 2.60 |
| | URPC (MedIA'22) | 85.11 | 75.63 | 6.72 | 1.51 |
| | SSNet (MICCAI'22) | 83.51 | 74.01 | 8.09 | 2.70 |
| | BCP (CVPR'23) | 84.64 | 75.56 | 7.18 | 2.30 |
| | ACMT (MedIA'23) | 84.34 | 75.04 | 7.02 | 2.33 |
| | DAE-MT (MedIA'24) | 81.74 | 71.94 | 8.10 | 1.78 |
| | MaskMatch (PR'25) | 84.38 | 74.63 | 6.84 | 1.45 |
| | **DuetMatch (Ours)** | **85.62** | **76.23** | **6.28** | **1.40** |

the consistency mask. This results in pseudo-labels that are both branch-specific and globally consistent:

$$PL_1^{cons} = PL_1 \odot P_{cons}, \qquad PL_2^{cons} = PL_2 \odot P_{cons}, \tag{20}$$

where $\odot$ denotes element-wise multiplication, and $PL_i^{cons}$ represents the refined pseudo-labels for the $i$th model branch.

Finally, the Cross-Guidance loss is reformulated using these consistency-enhanced pseudo-labels:

$$\mathcal{L}_{cm}^{cons} = l_P(P_1^{cm}, PL_2^{cons}) + l_P(P_2^{cm}, PL_1^{cons}). \tag{21}$$

By integrating the stable predictions from the frozen network components, this refinement strategy helps reduce noise in the pseudo-labels and improves the reliability of inter-branch supervision, ultimately contributing to more robust semi-supervised learning.

*3.6. Overall loss and inference*

In summary, the final training objective of our framework integrates supervised learning, duet-based self-supervision, and consistency-regularized cross-guidance. It is defined as follows:

$$\mathcal{L} = \mathcal{L}_{sup} + \mathcal{L}_{duet}^{dropout} + \beta \cdot \mathcal{L}_{cm}^{cons}, \tag{22}$$

where $\mathcal{L}_{sup}$ denotes the supervised loss on labeled data, $\mathcal{L}_{duet}^{dropout}$ is the Duet loss based on decoupled dropout perturbation, and $\mathcal{L}_{cm}^{cons}$ represents the cross-guidance loss with consistency-enhanced pseudo-labels. The hyperparameter $\beta$, which balances the contribution of cross-guidance loss, is empirically set to 0.5.

To maintain stable learning dynamics and promote temporal consistency, we update the frozen components of each branch using an exponential moving average (EMA) mechanism. Specifically:

$$\theta_1^{enc} \leftarrow w \cdot \theta_1^{enc} + (1-w) \cdot \theta_2^{enc}, \tag{23}$$

$$\theta_2^{dec} \leftarrow w \cdot \theta_2^{dec} + (1-w) \cdot \theta_1^{dec}, \tag{24}$$

where $w$ is the EMA decay factor, fixed at 0.99 throughout training. This strategy enables the frozen components to gradually evolve into more stable and representative versions of their corresponding counterparts.

During inference, we utilize the most consistent and generalized path through the network to generate the final prediction. In particular, both the frozen encoder and decoder are connected to produce the output mask, ensuring that the final segmentation benefits from the accumulated stability and representational power of the EMA-updated components.

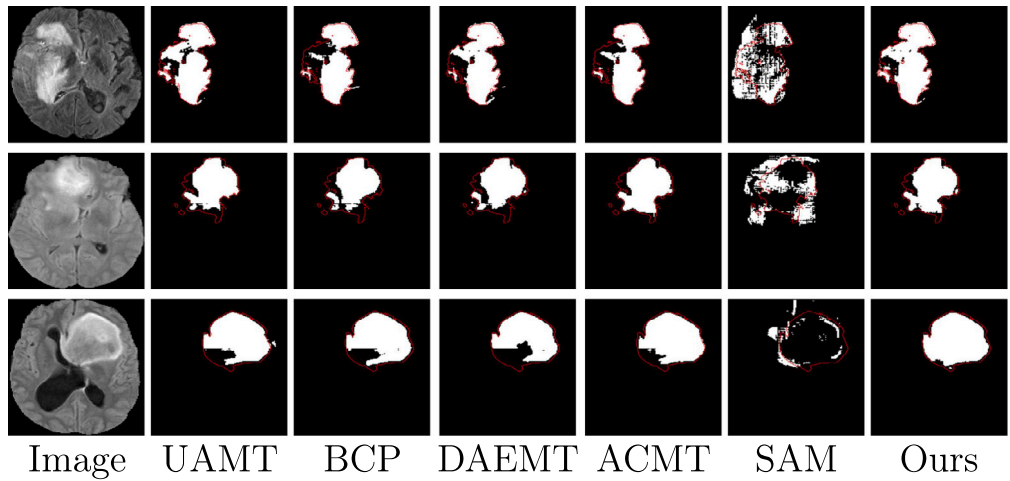

**Fig. 2.** Visualization of different methods on BraTS 2018 dataset. The red line represents the boundary of the ground truth mask.

## 4. Experiments

### 4.1. Datasets

We conducted experiments on four brain imaging datasets: BraTS2017, BraTS2018, BraTS2019, and ISLES2022 (Menze et al., 2015; Bakas et al., 2017, 2019; Hernandez Petzsche et al., 2022). The BraTS datasets, developed for the MICCAI brain tumor segmentation challenge, contain 285, 285, and 335 labeled cases respectively, categorized into high-grade and low-grade gliomas. Each case includes 3D MRI scans from four modalities (T1, T2, FLAIR, T1c); we primarily used FLAIR. Preprocessing involved brain region cropping and intensity normalization. Dataset splits were 200/25/60 as train/val/test for BraTS2017 and BraTS2018, and 250/25/60 for BraTS2019. All experiments were conducted independently with models trained from scratch.

ISLES2022 focuses on stroke lesion segmentation in 3D multimodal MRI, with 250 cases including DWI, ADC, and FLAIR. We used DWI and split the dataset into 150 training, 40 validation, and 60 testing samples.

### 4.2. Experimental settings

We used V-Net (Milletari et al., 2016) as the primary backbone in our experiments. The maximum number of iterations was set to 1000 for pretraining and 6000 for the main training phase. The batch size for both labeled and unlabeled data was fixed at 4. We employed the SGD optimizer with a learning rate of 0.01, momentum of 0.9, and a weight decay of $10^{-4}$. Data augmentation included random rotation, flipping, and random cropping with a patch size of $96 \times 96 \times 96$. The dropout ratio was set to 0.6. For consistency-based methods, the exponential moving average (EMA) decay was set to 0.99. All experiments were conducted on a single NVIDIA RTX A6000 GPU.

### 4.3. Results

**Quantitative Results.** Tables 1–4 report results on BraTS2017, BraTS-2018, BraTS2018, and ISLES2022 datasets with limited labeled data (4% or 10%). On BraTS2019, our method outperformed most competitors in Dice, Jaccard, and 95HD, with a notable lead under the 4% setting, despite slightly lower ASD. On BraTS2018, it consistently surpassed all baselines, achieving nearly 2% higher Dice than the runner-up. It also achieved top performance on all metrics for BraTS2017, highlighting strong robustness and generalization. For ISLES2022 with 10% labeled data, our method significantly outperformed DAE-MT and was only 1.6% lower than the fully supervised model. Moreover, the relatively lower performance of general models such as SAM and MedSAM may stem from the characteristics of the input images. Brain MRI scans in the BraTS and ISLES datasets often exhibit low contrast and poorly defined boundaries, while some tumors or stroke lesions are extremely small, making accurate segmentation challenging.

**Table 2**
Comparison of different methods on the FLAIR domain of BraTS 2018.

| | Method | DC (%) | JC (%) | 95HD (%) | ASD (%) |
|---|---|---|---|---|---|
| 100% labeled | V-Net | 84.13 | 74.46 | 9.83 | 3.14 |
| Zero-shot | SAM | 40.40 | 26.35 | 32.99 | 10.22 |
| | MedSAM | 25.74 | 15.93 | 27.49 | 8.46 |
| 10% labeled (20/200) | V-Net | 80.68 | 69.96 | 10.97 | 3.59 |
| | UA-MT (MICCAI'19) | 84.31 | 74.95 | 8.85 | 2.76 |
| | SASSNet (MICCAI'20) | 84.85 | 75.69 | 8.55 | 2.67 |
| | DTC (AAAI'21) | 84.25 | 74.88 | 9.21 | 2.85 |
| | MC-Net (MICCAI'21) | 84.78 | 75.08 | 8.11 | 1.57 |
| | URPC (MedIA'22) | 85.40 | 76.59 | 7.63 | 2.00 |
| | SSNet (MICCAI'22) | 84.63 | 74.83 | 7.75 | 1.52 |
| | BCP (CVPR'23) | 85.57 | 76.06 | 6.86 | 1.43 |
| | ACMT (MedIA'23) | 83.43 | 74.28 | 10.18 | 3.65 |
| | DAE-MT (MedIA'24) | 84.41 | 74.81 | 7.96 | 1.80 |
| | MaskMatch (PR'25) | 85.31 | 76.14 | 8.19 | 2.47 |
| | DuetMatch (Ours) | **87.60** | **78.67** | **6.31** | **1.33** |

**Qualitative Results.** Figs. 2, 3, 4 present visual comparisons of the predicted segmentation results from our method and other baselines on the middle slice of the BraTS2018, BraTS2019, and ISLES2022 datasets. Our method demonstrates greater robustness and better coverage of the ground truth regions, particularly in areas where other methods fail to identify lesions and incorrectly classify them as background. Furthermore, our approach yields more accurate boundaries and preserves the overall shape of the target structures more effectively than competing methods.

### 4.4. Ablation studies

**Feature Loss.** We evaluated three loss functions, including MSE (L2), MAE (L1), and cosine similarity, for feature-level consistency between teacher and student encoders on the 4% labeled BraTS2019 dataset (Table 5). MSE and MAE both performed well overall; MAE yielded the best 95HD but was slightly behind MSE on Dice, Jaccard, and ASD. Due to its stronger overall performance, MSE was chosen as the default. Despite its popularity, cosine similarity consistently underperformed across all metrics, indicating MSE provides a more robust and balanced signal for consistency training.

**Table 3**
Comparison of different methods with varying labeled data ratios on the FLAIR domain of BraTS 2019.

| | Method | Dice (%) | Jaccard (%) | 95HD (%) | ASD (%) |
|---|---|---|---|---|---|
| 100% labeled | V-Net | 84.38 | 75.77 | 11.11 | 4.32 |
| Zero-shot | SAM | 38.73 | 25.18 | 47.54 | 18.84 |
| | MedSAM | 37.38 | 24.17 | 36.95 | 15.03 |
| 4% labeled (10/250) | V-Net | 75.40 | 66.28 | 16.46 | 7.97 |
| | UA-MT (MICCAI'19) | 82.18 | 72.90 | 12.11 | 4.17 |
| | SASS-Net (MICCAI'20) | 83.30 | 73.87 | 9.24 | 1.64 |
| | DTC (AAAI'21) | 82.07 | 73.00 | 10.36 | 2.93 |
| | MC-Net (MICCAI'21) | 83.44 | 74.49 | 9.34 | **1.18** |
| | URPC (MedIA'22) | 78.51 | 69.78 | 15.14 | 7.20 |
| | SSNet (MICCAI'22) | 81.32 | 72.03 | 12.89 | 4.14 |
| | BCP (CVPR'23) | 83.84 | 75.12 | 10.51 | 3.88 |
| | ACMT (MedIA'23) | 82.33 | 73.01 | 10.85 | 4.04 |
| | DAE-MT (MedIA'24) | 82.07 | 72.91 | 11.56 | 2.44 |
| | MaskMatch (PR'25) | 81.65 | 72.41 | 11.27 | 2.52 |
| | **DuetMatch (Ours)** | **86.23** | **77.92** | **6.63** | 1.35 |
| 10% labeled (25/250) | V-Net | 82.38 | 72.37 | 10.31 | 2.91 |
| | UA-MT (MICCAI'19) | 85.38 | 76.84 | 9.74 | 2.45 |
| | SASS-Net (MICCAI'20) | 84.96 | 76.74 | 8.46 | 2.35 |
| | DTC (AAAI'21) | 85.48 | 76.89 | 7.98 | **1.17** |
| | MC-Net (MICCAI'21) | 85.66 | 77.14 | 9.76 | 2.31 |
| | URPC (MedIA'22) | 84.81 | 76.47 | 9.41 | 2.52 |
| | SSNet (MICCAI'22) | 85.02 | 76.41 | 9.81 | 2.53 |
| | BCP (CVPR'23) | 85.78 | 77.30 | 9.35 | 2.33 |
| | ACMT (MedIA'23) | 86.26 | 77.80 | 8.56 | 2.39 |
| | DAE-MT (MedIA'24) | 84.51 | 75.62 | 8.32 | 1.44 |
| | MaskMatch (PR'25) | 83.83 | 74.81 | 8.56 | 1.26 |
| | **DuetMatch (Ours)** | **86.77** | **78.62** | 8.06 | 2.14 |

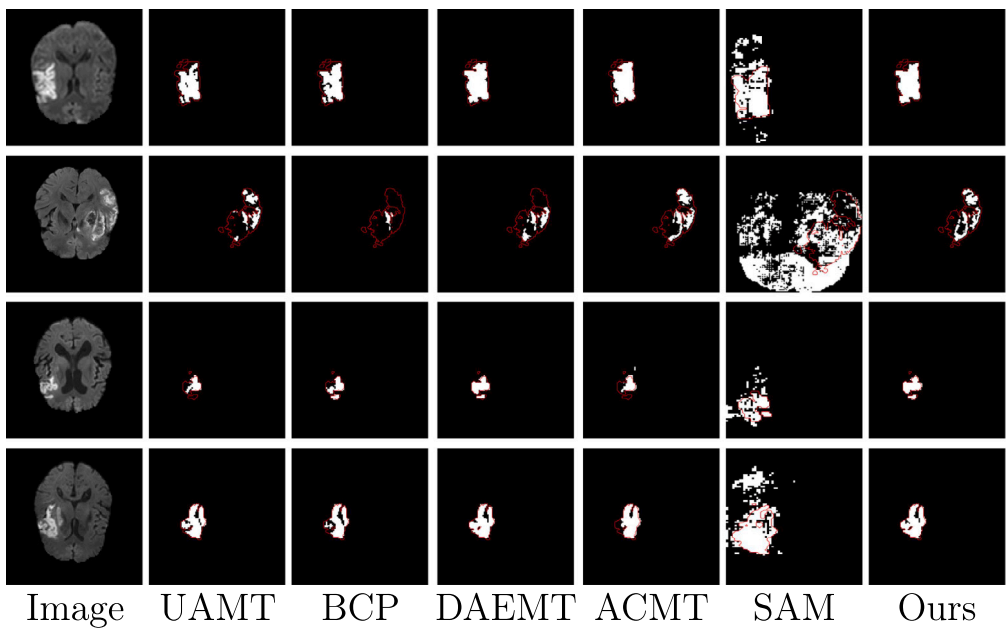

**Fig. 3.** Visualization of different methods on ISLES 2022.

**Ratio Dropout Selection.** We investigated the impact of dropout ratios (0.0 to 0.8) on model performance using 4% labeled data from the BraTS2019 dataset. As shown in Fig. 5, validation metrics (Dice, Jaccard) improved with higher dropout, indicating better generalization. However, test results showed that a 0.6 ratio achieved the best overall performance, balancing regularization and learning capacity. Although 0.8 gave the highest validation scores, it slightly underperformed on the test set, suggesting overfitting. Thus, we selected 0.6 as the optimal dropout ratio for our framework.

**Component Analysis.** We conducted an ablation study using K-Fold Cross Validation to assess each component of our framework (Table 6). Starting from a baseline Dice score of 82.54%, adding Decoupled Dropout Perturbation (DDP) improved generalization, raising Dice to 83.94% and reducing ASD. Incorporating Pairwise CutMix Cross-Guidance (PCMCG) further boosted performance to 85.00% Dice. Combining DDP and PCMCG provided additional gains in Dice and 95HD. Finally, adding the Consistency Matching strategy achieved the best results (86.05% Dice and a 95HD of 7.99) by filtering noisy pseudo-labels and enhancing training stability. Overall, each component contributed to improved segmentation accuracy and robustness.

**Training time & GPU usage.** As shown in Table 7, our method requires a longer training time than other approaches, mainly due to device may lead to longer training durations. Despite the increased complexity, our method maintains a moderate level of GPU utilization, with consumption remaining close to the average across all compared methods.

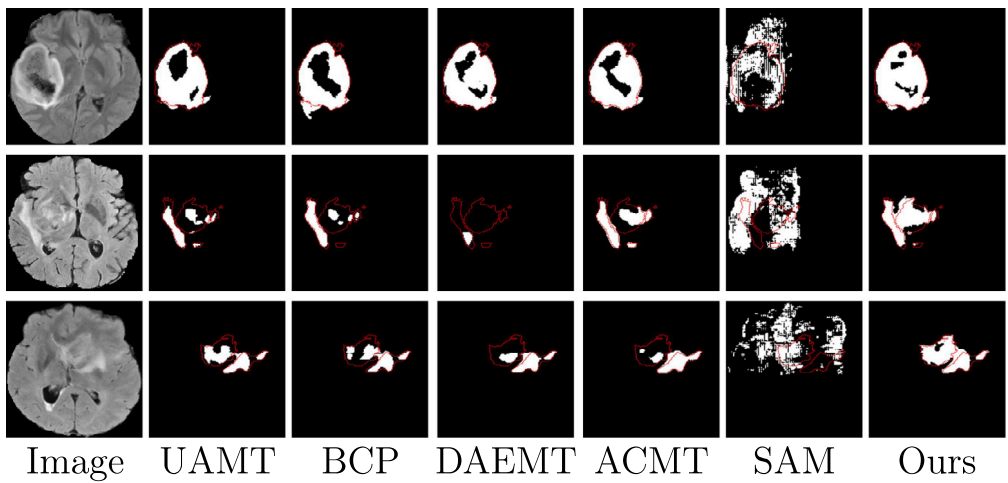

Fig. 4. Visualization of different methods on BraTS 2019.

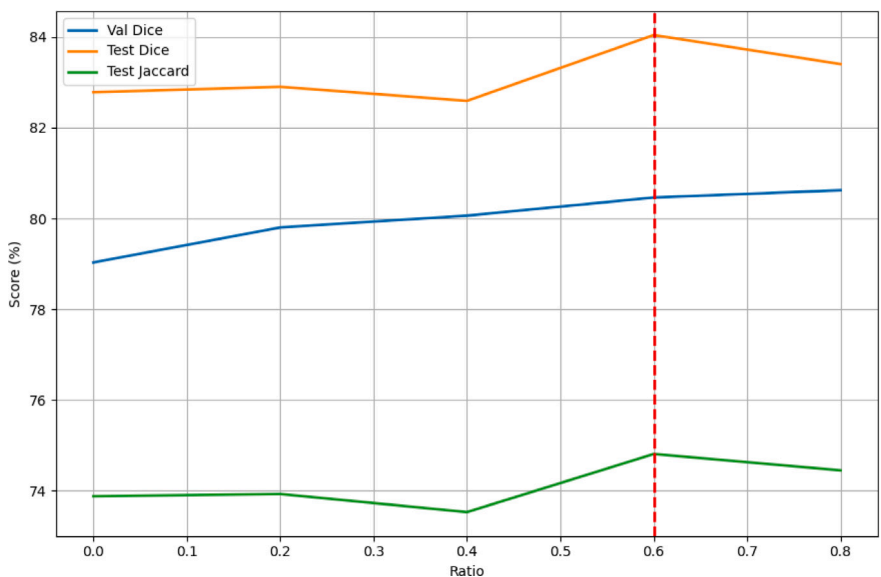

Fig. 5. Demonstration of different dropout ratios on model performance.

Table 4
Comparison of different methods on the DWI domain of ISLES 2022.

| | Method | DC (%) | JC (%) | 95HD (%) | ASD (%) |
|---|---|---|---|---|---|
| 100% labeled | V-Net | 53.27 | 42.34 | 18.71 | 3.88 |
| Zero-shot | SAM | 35.55 | 23.43 | 13.75 | 5.15 |
| | MedSAM | 19.29 | 11.94 | 17.48 | 7.22 |
| 10% labeled (15/150) | V-Net | 36.93 | 27.70 | 22.66 | 10.22 |
| | UA-MT (MICCAI'19) | 41.83 | 31.27 | 21.04 | 4.83 |
| | SASSNet (MICCAI'20) | 40.87 | 31.19 | 21.41 | 3.66 |
| | DTC (AAAI'21) | 39.27 | 29.24 | 21.96 | 4.26 |
| | MC-Net (MICCAI'21) | 41.82 | 31.60 | 20.95 | **2.34** |
| | URPC (MedIA'22) | 43.62 | 32.40 | 20.34 | 3.89 |
| | SSNet (MICCAI'22) | 42.39 | 32.28 | 20.55 | 3.49 |
| | BCP (CVPR'23) | 43.84 | 33.51 | 20.61 | 3.61 |
| | ACMT (MedIA'23) | 45.08 | 34.28 | 18.05 | 3.21 |
| | DAE-MT (MedIA'24) | 47.98 | 37.32 | 20.51 | 5.21 |
| | DuetMatch (Ours) | **51.67** | **40.62** | 19.96 | 5.43 |

Table 5
Comparison of different losses for feature consistency.

| Loss | DC (%) | JC (%) | 95HD (%) | ASD (%) |
|---|---|---|---|---|
| Cosine | 82.01 | 72.95 | 10.07 | 2.68 |
| MAE | 82.68 | 73.73 | **9.51** | 2.65 |
| MSE | **82.78** | **73.88** | 9.69 | **2.65** |

## 5. Discussion and concluding remarks

In this work, we propose a dual-branch framework with asynchronous optimization objectives, named DuetMatch, which integrates Decoupled Dropout Perturbation to enhance the model's consistency and generalization capabilities. Furthermore, our approach incorporates Pairwise CutMix Cross-Guidance and Consistency Matching to improve model diversity and facilitate more effective learning from unlabeled data. Experimental results on multiple brain segmentation benchmarks demonstrate the effectiveness and robustness of our method the additional functional complexity introduced to the baseline. However, it is important to note that training time analysis is relative, as it depends on the hardware configuration and computational conditions during training. For instance, concurrent processes running on the same across various scenarios. Although the absolute Dice gains over strong baselines like BCP and URPC may seem modest, they are consistent across all datasets and metrics, which is meaningful under low-label semi-supervised settings. Such steady improvements reflect enhanced convergence stability and boundary precision rather than random variation. Moreover, DuetMatch achieves these results without added inference cost, as the dual-branch structure mainly benefits training dynamics. From a practical standpoint, this stability and reliability are often more valuable than large but inconsistent numeric gains.

**Table 6**
Comparison of different components using K-Fold Cross Validation on 4% labeled settings of BraTS2019.

| Baseline | DDP | PCMCG | Consistency matching | Dice (%) | Jaccard (%) | 95HD (%) | ASD (%) |
|---|---|---|---|---|---|---|---|
| ✓ | | | | 82.54 ± 2.14 | 73.49 ± 2.85 | 10.85 ± 2.33 | 2.78 ± 1.60 |
| ✓ | ✓ | | | 83.94 ± 0.48 | 75.27 ± 0.67 | 9.00 ± 0.74 | **1.60** ± 0.62 |
| ✓ | | ✓ | | 85.00 ± 1.03 | 76.43 ± 1.25 | 9.40 ± 0.87 | 2.85 ± 0.79 |
| ✓ | ✓ | ✓ | | 85.13 ± 0.20 | 76.85 ± 0.24 | 8.60 ± 0.51 | 2.54 ± 0.25 |
| ✓ | ✓ | ✓ | ✓ | **86.05** ± 0.99 | **77.76** ± 0.96 | **7.99** ± 0.44 | 2.13 ± 0.74 |

**Table 7**
Additional comparison of training time and GPU usage followed by Table 1.

| Method | Training time | GPU usage |
|---|---|---|
| Sup-Only | ≈0.5 h | 4 GB |
| UA-MT | ≈2.5 h | 12.7 GB |
| SASS-Net | ≈4 h | 8.5 GB |
| DTC | ≈3 h | 7.1 GB |
| MC-Net | ≈2.5 h | 15.3 GB |
| URPC | ≈2 h | 15.6 GB |
| SSNet | ≈3 h | 20.0 GB |
| BCP | ≈3.5 h | 7.6 GB |
| ACMT | ≈7.5 h | 8.8 GB |
| MaskMatch | ≈3 h | 18.2 GB |
| DAE-MT | ≈7.5 h | 10.1 GB |
| **DuetMatch** | **≈7.5 h** | **13.9 GB** |

Future work could explore adaptive pseudo-label refinement strategies, such as other uncertainty estimation or outlier rejection methods, to further reduce noise in challenging cases. Integrating lightweight architectures or dynamic branch pruning could optimize computational efficiency without sacrificing performance. Expanding validation to diverse imaging modalities and multicenter datasets would strengthen the method's robustness and clinical applicability. Finally, investigating hybrid approaches that combine *DuetMatch* with domain adaptation techniques could enhance its utility for underrepresented or imbalanced data distributions. These advancements would solidify the framework's role in broader medical image analysis.

**CRediT authorship contribution statement**

**Thanh-Huy Nguyen:** Writing – original draft, Validation, Supervision, Methodology, Conceptualization. **Hoang-Thien Nguyen:** Writing – original draft, Visualization, Validation, Methodology, Conceptualization. **Vi Vu:** Writing – review & editing, Writing – original draft, Visualization, Validation. **Ba-Thinh Lam:** Writing – review & editing, Validation, Methodology. **Phat Huynh:** Writing – review & editing, Software, Resources, Investigation. **Tianyang Wang:** Writing – review & editing, Supervision, Resources. **Xingjian Li:** Writing – review & editing, Supervision, Project administration, Conceptualization. **Ulas Bagci:** Writing – review & editing, Software, Resources, Project administration. **Min Xu:** Writing – review & editing, Writing – original draft, Supervision, Software, Resources, Project administration, Investigation, Funding acquisition.

**Declaration of competing interest**

The authors declare that they have no known competing financial interests or personal relationships that could have appeared to influence the work reported in this paper.

**Acknowledgments**

This work was supported in part by U.S. NSF grants DBI-2238093. We thank AI VIETNAM for supporting us with GPUs to conduct experiments.

**Data availability**

Data will be made available on request.